\documentclass[conference]{IEEEtran}
\IEEEoverridecommandlockouts
\usepackage{cite}
\usepackage{amsmath,amssymb,amsfonts}
\usepackage{algorithmic}
\usepackage{graphicx}
\usepackage{float}
\usepackage{textcomp}
\usepackage{subcaption}
\usepackage{multirow}
\usepackage{balance}
\usepackage{xcolor}
\usepackage{tabularx}
\def\BibTeX{{\rm B\kern-.05em{\sc i\kern-.025em b}\kern-.08em
    T\kern-.1667em\lower.7ex\hbox{E}\kern-.125emX}}

\definecolor{lightred}{HTML}{FBDEDE}
\definecolor{lightyellow}{HTML}{FBF7EF}

\makeatletter
\newcommand{\linebreakand}{%
  \end{@IEEEauthorhalign}
  \hfill\mbox{}\par
  \mbox{}\hfill\begin{@IEEEauthorhalign}
}
\makeatother

\begin{document}

\title{MEX: Memory-efficient Approach to \\ Referring Multi-Object Tracking}

\author{
\IEEEauthorblockN{Huu-Thien Tran}
\IEEEauthorblockA{\textit{University of Science} \\
\textit{Vietnam National University, Ho Chi Minh City}\\
Ho Chi Minh City, Vietnam \\
20120584@student.hcmus.edu.vn}
\and
\IEEEauthorblockN{Phuoc-Sang Pham}
\IEEEauthorblockA{\textit{University of Science} \\
\textit{Vietnam National University, Ho Chi Minh City}\\
Ho Chi Minh City, Vietnam \\
20120364@student.hcmus.edu.vn}
\and

\linebreakand
\IEEEauthorblockN{Thai-Son Tran}
\IEEEauthorblockA{\textit{University of Science} \\
\textit{Vietnam National University, Ho Chi Minh City}\\
Ho Chi Minh City, Vietnam \\
ttson@fit.hcmus.edu.vn}
\and
\IEEEauthorblockN{Khoa Luu}
\IEEEauthorblockA{\textit{Department of Computer Science and Computer Engineering} \\
\textit{University of Arkansas}\\
Fayetteville, USA \\
khoaluu@uark.edu}
}

\maketitle

\begin{abstract}
Referring Multi-Object Tracking (RMOT) is a relatively new concept that has rapidly gained traction as a promising research direction at the intersection of computer vision and natural language processing. Unlike traditional multi-object tracking, RMOT identifies and tracks objects and incorporates textual descriptions for object class names, making the approach more intuitive. Various techniques have been proposed to address this challenging problem; however, most require the training of the entire network due to their end-to-end nature. Among these methods, iKUN has emerged as a particularly promising solution. Therefore, we further explore its pipeline and enhance its performance. In this paper, we introduce a practical module dubbed \textbf{M}emory-\textbf{E}fficient Cross-modality -- \textbf{MEX}. This memory-efficient technique can be directly applied to off-the-shelf trackers like iKUN, resulting in significant architectural improvements. Our method proves effective during inference on a single GPU with 4 GB of memory. Among the various benchmarks, the Refer-KITTI dataset, which offers diverse autonomous driving scenes with relevant language expressions, is particularly useful for studying this problem. Empirically, our method demonstrates effectiveness and efficiency regarding HOTA tracking scores, substantially improving memory allocation and processing speed.
\end{abstract}


\section{Introduction}

The traditional multi-object tracking (MOT) task focuses on tracking specific classes of objects in each video frame, playing a crucial role in video understanding. Despite substantial developments and breakthroughs in this field, the inherent specificity of class names limits their flexibility and generalization. To address this issue, an emerging task named referring multi-object tracking (RMOT) integrates the multi-object tracker with additional language expressions as semantic cues, frame-by-frame. This procedure is more general and flexible as predictions per frame can be arbitrary, targeting only the referred objects. For example, with the language description ``moving cars in black,'' the tracker will predict trajectories corresponding only to the specified prompt, ignoring parked cars or cars of other colors. While this method enhances flexibility, it adds complexity to the tracking pipeline, which already involves detection, association, and the additional referring task.

There are numerous approaches to solving this problem. TransRMOT \cite{wu2023referring}, an extension of MOTR \cite{zeng2021motr}, incorporates a linguistic module. MENDER \cite{nguyen2023type} presents a unified network that addresses three distinct components within the cross-modality module, including the prompt as a semantic cue and visual data sources from the frame and corresponding trajectories over time. These networks are all trained end-to-end and have made significant contributions to the field. However, they face limitations when applied to datasets with different distributions, necessitating a complete retraining of the network. Recently, iKUN \cite{du2024ikun} has approached this task more effectively. It decouples the problem into two sub-tasks, termed \textit{tracking-to-referring}. It allows the tracker to remain frozen during training, while the referring module can be used plug-and-play with any tracker.

\begin{figure}
    \centering
    \includegraphics[width=\linewidth]{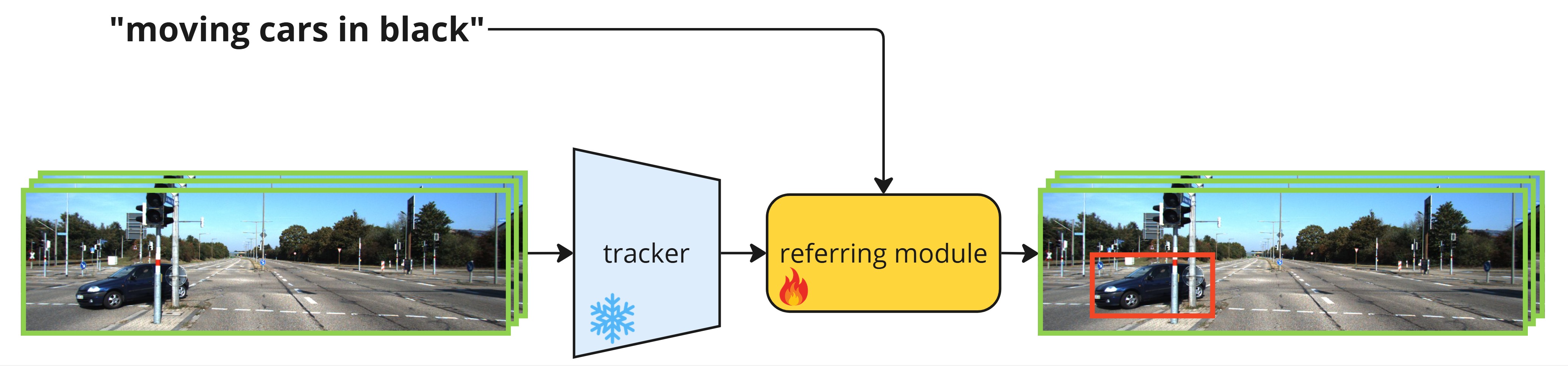}
    \caption{The plug-and-play design of referring module introduced by iKUN \cite{du2024ikun}. Best viewed in color.}
    \label{fig:rmot_plugnplay}
\end{figure}

\begin{figure*}[h]
      \centering
      \includegraphics[width=\textwidth]{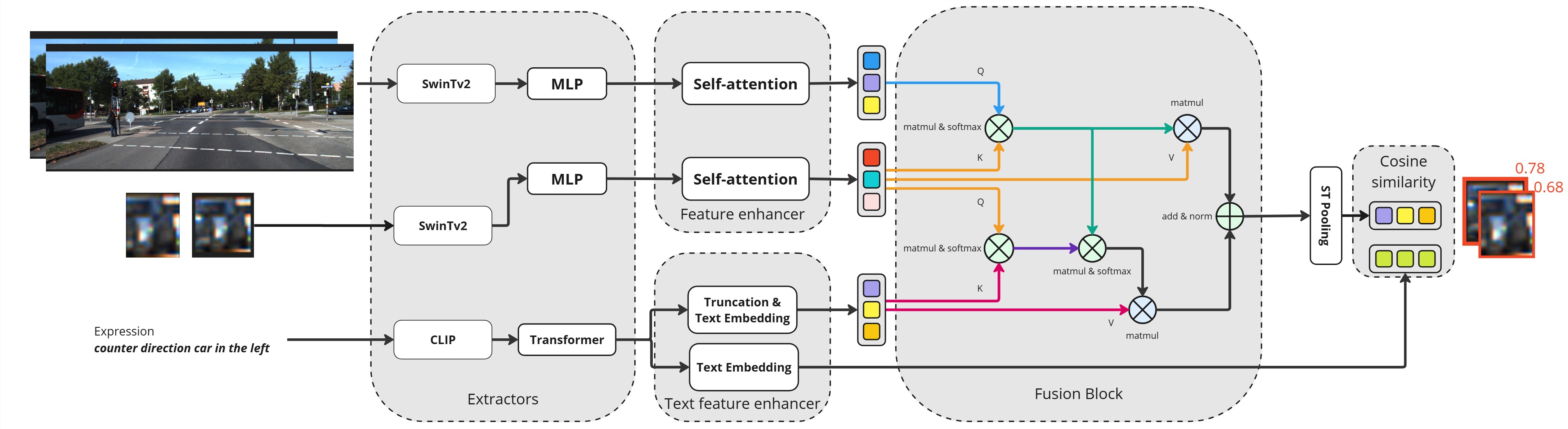}
      \caption{
        \textbf{Overview of our MEX.}
        This is our designated pipeline for the referring module, which can be attached at the end of the tracking process to address the referring multi-object tracking task. We employ SwinV2-T \cite{liu2022swin} for visual embedding and the CLIP \cite{radford2021clip} text encoder for textual embedding. After the truncation step, the truncated embedded features are fed into our designed fusion block. The output of the fusion block is then passed through a spatio-temporal (ST) pooling layer, followed by a cosine similarity criterion between it and the original embedded features. Best viewed in color.
      }
      \label{fig:overall}
\end{figure*}

With that motivation, our method focuses on improving effectiveness during training and inference processes. Although the iKUN pipeline aligns well with the problem's purpose, there is still room for improvements in memory efficiency. To address this, we incorporate an elegant referring module using our designated MEX mechanism further to enhance the overall performance of the tracking pipeline. We conduct extensive experiments on the recently released Refer-KITTI dataset, and our method achieves compelling results compared to iKUN. Specifically, our method not only surpasses the predecessor with a \textbf{0.5\%} increase in the HOTA~\cite{luiten2021hota} score but also demonstrates improved efficacy in terms of memory usage and inference speed, achieving approximately \textbf{0.5x} lower memory consumption and \textbf{1.5x} faster inference speed.

\section{Related Work}

{\bfseries Multi-Object Tracking.}
Prevailing approaches to multi-object tracking problems can be classified into two main paradigms: tracking-by-detection and tracking-by-attention. Tracking-by-detection involves two consecutive stages: using a detector to predict the bounding boxes of objects and their corresponding features, followed by an association step to match these instances across frames. SORT \cite{wojke2017simple} employs the Kalman filter for motion modeling and associates tracking instances based on the intersection-over-union (IoU) of bounding boxes. DeepSORT \cite{Wojke2018deepsort} enhances this method by integrating a deep learning-based module to extract the appearance features of objects. More recently, ByteTrack \cite{zhang2022bytetrack} treats detection boxes as bytes and associates all of them, significantly boosting the tracker's performance. Advanced techniques in BoT-SORT \cite{aharon2022botsort}, and OC-SORT \cite{cao2023observation} further refine the association and post-processing steps, achieving even better performance. On the other hand, the advent of the Transformer architecture \cite{vaswani2017attention} has driven considerable advancements in the tracking-by-attention paradigm in recent years. TrackFormer \cite{meinhardt2021trackformer} introduces tracking instances as queries during the detection step. MOTR \cite{zeng2021motr} expands on this by incorporating a query interaction module, and MOTRv2 \cite{zhang2023motrv2} further improves its performance by initializing queries using detection results from YOLOX \cite{yolox2021}. Furthermore, by integrating historical features across frames as memory, MeMOT \cite{cai2022memot}, and MeMOTR \cite{MeMOTR} noticeably improve the performance of attention-based trackers.

{\bfseries Referring Multi-Object Tracking.}
Referring multi-object tracking is an emerging challenge, whereas referring single-object tracking, or segmentation, has been studied for years and has achieved good performance results \cite{ZHAO202310, zhou2023joint, zheng2023towards, wu2022language, botach2021end}. However, due to some intrinsic limitations, referring multi-object tracking cannot be directly expanded from these approaches. Precedent methods employ various techniques to tackle this additional textual referring expression. MENDER \cite{nguyen2023type} utilizes the cross-modality fusion module to handle three distinct inputs, including features from the video frame, tracking instances, and textual captions. To further enhance performance, they further implement third-order tensor decomposition. TransRMOT \cite{wu2023referring} places the cross-attention module at the beginning of the pipeline, aggregating the textual features into the detection step. Despite the flexibility gained, training these end-to-end tracking networks entirely for diverse video distributions incurs considerable computational costs due to the demanding training process. Recently, iKUN \cite{du2024ikun} designed a plug-and-play pipeline for referring multi-object tracking problems by decoupling the task into two sub-tasks: tracking and referring. It allows the tracker network to remain frozen during training, offering flexibility in utilizing different off-the-shelf trackers. Inspired by this approach, we explore this paradigm further and facilitate the method with our memory-efficient technique during the referring task, enhancing the performance of the pipeline.

\section{Methodology}

\begin{figure*}[tp!]
    \centering
    \subfloat[iKUN's cascade attention]{
        \includegraphics[width=.3\textwidth]{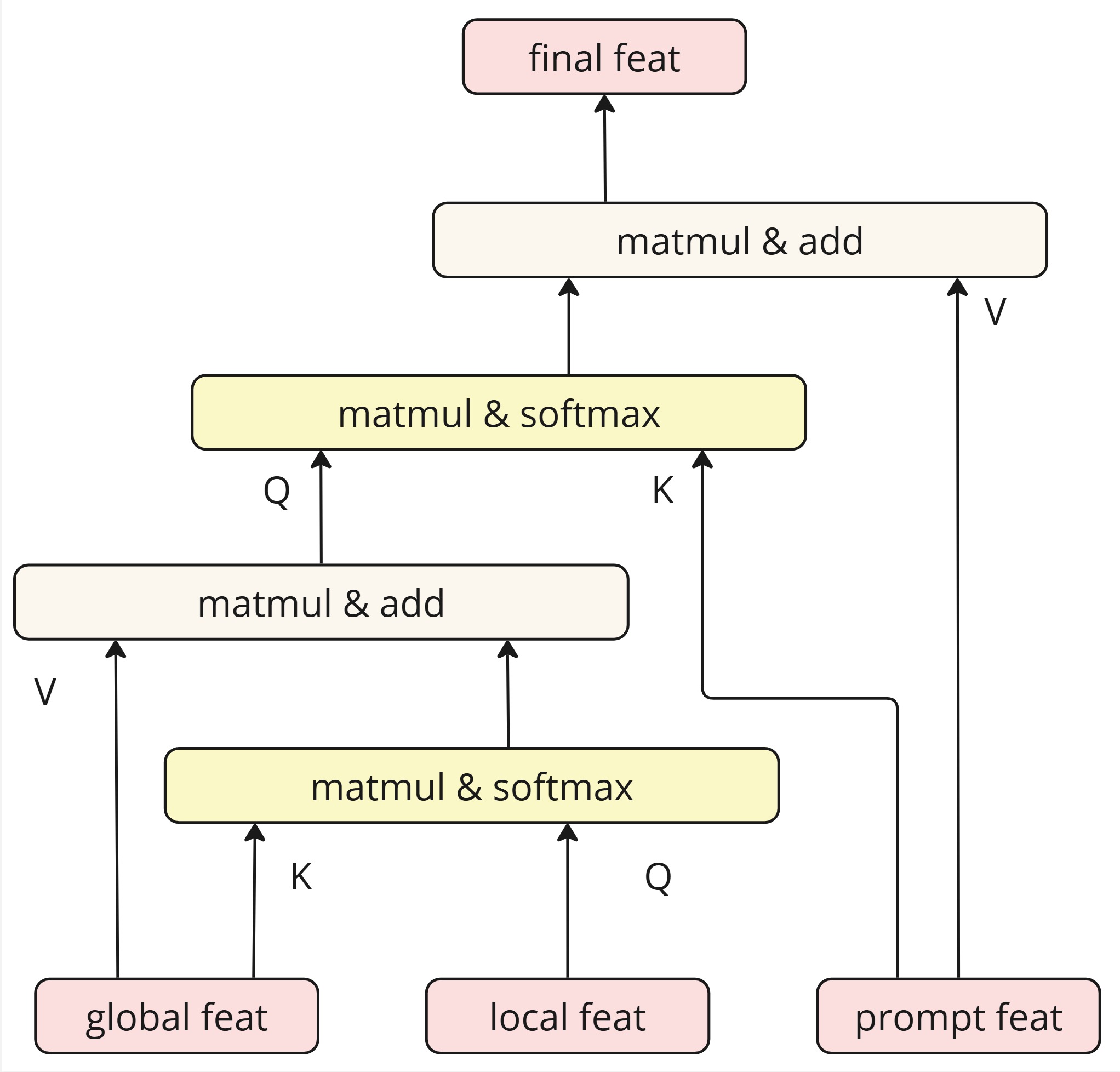}
    }
    \subfloat[MEX attention]{
        \includegraphics[width=.3\textwidth]{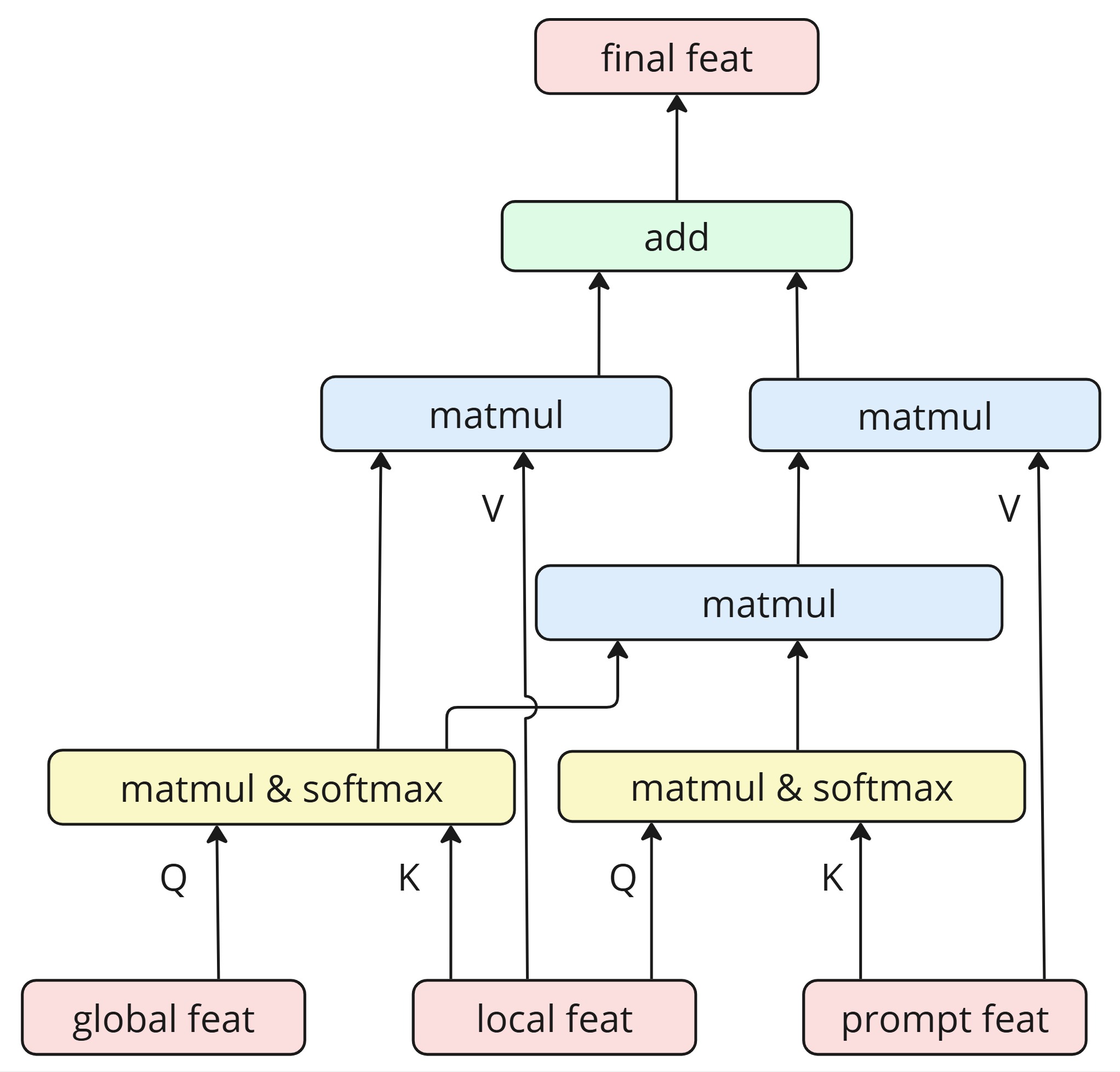}
    }
    \caption{
    \textbf{Comparison between previous and our design for fusion block.}
    (a) Cascade attention architecture introduced in iKUN \cite{du2024ikun}, which has the best performance.
    (b) Our MEX attention.
    The \colorbox{lightred!70}{global feat}, \colorbox{lightred!70}{local feat}, and \colorbox{lightred!70}{prompt feat} are the different appellations for features from the video frame, boxes of tracking instances, and the language expression, respectively. Note that before becoming query, key, and value vectors, the feature vectors are projected into desired dimensions using linear layers or repeating along a specific dimension. Also, the add operation in the \colorbox{lightyellow}{matmul \& add} block means taking the result from matrix multiplication plus the query vector. Due to the purpose of visualization, these notable steps are omitted here. Best viewed in color.}
    \label{fig:fusion}
\end{figure*}

%
In this section, we first discussed the preliminary related to our problem, including the detailed formulation of the problem and the materials coming from \cite{du2024ikun} as the baseline method we want to improve in \ref{method:pre}. Then, we introduce our fusion block using scaled dot-product cross-modality attention in \ref{method:net}.

\subsection{Preliminary}
\label{method:pre}
As described in \cite{du2024ikun}, the input of RMOT consists of a sequence $\mathcal{I} = \left\{ I_t\right\}^K_{k=1}$ with $K$ frames and $\mathcal{P} = \left\{ P_\ell\right\}^L_{\ell=1}$ as a language expression with $L$ tokens. Given a tracker $\mathcal{F}_{tr}(\cdot)$, all predicted trajectories are $\mathcal{T} = \mathcal{F}_{tr}(\mathcal{I})$, where $\mathcal{T} = \left\{ \mathcal{T}_1, \mathcal{T}_2, \cdots, \mathcal{T}_N \right\}$ includes $N$ tracking instances. This tracking module is followed by a referring module $\mathcal{F}_{ref}(\cdot)$, applied to mark candidate instances $\mathcal{T}$ by the textual prompt with scoring criterion as $\mathcal{S} = \mathcal{F}_{ref}(\mathcal{I}, \mathcal{T}, \mathcal{P})$. In the iKUN's pipeline, the similarity calibration module further refined this referring score, which can enhance the overall performance. In that case, we also integrate this module into our pipeline and run experiments to observe the results. Finally, candidates $\mathcal{T}$ are filtered by the refined scores $\mathcal{S}^\prime$ and the tracking outputs $\mathcal{T}^\prime$, with $M$ created trajectories ($M \leq N$).

The empirical effectiveness of utilizing three components, $\mathcal{I}$, $\mathcal{T}$, and $\mathcal{P}$, stems from their enhanced spatial-temporal relationship between frames, trajectories, and linguistic expressions. Designing a suitable architecture for these inputs is particularly challenging due to the inherent complexity of the RMOT task. Specifically, each prompt can address multiple objects in a frame, while multiple prompts can target a single object. Existing methods \cite{nguyen2023type,du2024ikun} also consider these three factors, with advanced techniques to deal with them. However, they are intrinsically intricate and challenging to replicate on resource-constrained devices such as personal computers. Further details are discussed in \ref{method:net}.

To elaborate on the techniques outlined in the original iKUN paper, we present a detailed exploration of the similarity calibration module. This module alleviates non-uniform and open test sets encountered in RMOT scenarios. Initially, \eqref{simcali} defines the normalized similarity between language descriptions during test-time, denoted as $w_{ij}$.

\begin{equation}
w_{ij} = \dfrac{\exp(\tau \cdot x_{ij})}{\sum_k{\exp(\tau \cdot x_{ik})}}
\label{simcali}
\end{equation}

where $\tau$ is the temperature parameter ($\tau$ is set to 100), and $x_{ij}$ represents the similarity score estimated by a language model. Also, the frequencies of all distinct $N_{tr}$ training language descriptions are estimated using the same language model as previously described, denoted as $\left\{p_i^{tr}\right\}_{i=1}^{N_{tr}}$. Subsequently, the pseudo frequency is computed as $p^{ts}_j = \sum_i {w_{ij} \cdot p^{tr}_i}$. Then, the referring score $s_j$ after being refined is given by $s_j^\prime = s_j + a \cdot p^{ts}_j + b$, where constants $a = 8$ and $b = -0.1$.

\subsection{Fusion Block}
\label{method:net}
We propose a novel fusion block that can tackle the inputs from the three sources above of data. To effectively resolve the interdependencies among these three components, we introduce a method called \textbf{Memory-Efficient Cross-Modality Attention} -- \textbf{MEXAttn}. Unlike the traditional scale dot-product attention described in \cite{vaswani2017attention}, which typically handles pairwise relationships between two sets of criteria, our technique can simultaneously accommodate three types of components. This architectural enhancement enables the transformer architecture to learn correlations more effectively. Our methodology is illustrated in Fig.~\ref{fig:fusion}, alongside a depiction of the cascade attention method introduced in \cite{du2024ikun}, which outperforms other designs in their knowledge unification module.

To formulate the MEX attention mechanism, we first need to look at the original scale dot-product attention algorithm described in 
\eqref{eq:attn}.

\begin{equation}
    \textrm{Attn}(Q,K,V) = \sigma\left( \dfrac{QK^\top}{\sqrt{d_k}}\right)V
    \label{eq:attn}
\end{equation}

where $Q$, $K$, and $V$ represent the query, key, and value vector, respectively ($K^\top$ denotes the transpose of $K$); $\sqrt{d_k}$ serves as a scaling factor, where $d_k$ is the dimension of the features; and $\sigma(\cdot)$ denotes the softmax function. Notably, $Q$ interacts exclusively with one data source, whereas $K$ and $V$ interact with the other. We extend this concept by inserting an additional matrix multiplication in (\ref{eq:attn}), creating a more elegant way to establish correlations between an extra pair of data sources. The architectural design of our referring module is portrayed in Fig.~\ref{fig:fusion}(b).

Here, we outline the calculation steps in detail. Initially, we correlate the frame's features $I_t$ and the features of tracking instances $\mathcal{T}_n$, treating them as query and key vectors, respectively, in \eqref{eq:attn_it}. Similarly, we apply this correlation process between $\mathcal{T}_n$ and the features of the expression $\mathcal{P}_j$ in \eqref{eq:attn_tp}. In order to simplify the problem, all the feature extractors and projection layers are collectively denoted as $f(\cdot)$. Subsequently, the output from these steps is represented as $p_{\mathcal{I}|\mathcal{T}}$ and $p_{\mathcal{T}|\mathcal{P}}$, followed by a matrix multiplication operation as \eqref{eq:attn_itp}. 

\begin{equation}
    p_{\mathcal{I}|\mathcal{T}} = \sigma\left( \dfrac{f(I_t)\cdot f(\mathcal{T}_n)^\top}{\sqrt{d_k}}\right)
    \label{eq:attn_it}
\end{equation}

\begin{equation}
    p_{\mathcal{T}|\mathcal{P}} = \sigma\left( \dfrac{f(\mathcal{T}_n)\cdot f(\mathcal{P}_j)^\top}{\sqrt{d_k}}\right)
    \label{eq:attn_tp}
\end{equation}

\begin{equation}
p_{\mathcal{I}|\mathcal{T},\mathcal{P}} = p_{\mathcal{I}|\mathcal{T}} \cdot p_{\mathcal{T}|\mathcal{P}}
\label{eq:attn_itp}
\end{equation}

At the end, the output of our MEX attention mechanism can be computed as \eqref{eq:remexattn}.

\begin{equation}
\begin{aligned}
    \textrm{MEXAttn}(\mathcal{I}, \mathcal{T}, \mathcal{P}) & = p_{\mathcal{I}|\mathcal{T}} \cdot f(\mathcal{T}_n) + p_{\mathcal{I}|\mathcal{T},\mathcal{P}} \cdot f(\mathcal{P}_j)
\end{aligned}
\label{eq:remexattn}
\end{equation}

\section{Experiments}

\subsection{Benchmark datasets}
\textbf{Refer-KITTI} \cite{wu2023referring} is a publicly available dataset designed for referring multi-object tracking, an extension of the KITTI \cite{Geiger2012CVPR}. Refer-KITTI consists of 18 high-resolution, lengthy videos containing 818 expressions, each corresponding to an average of 10.7 objects. This dataset covers various scenes, such as pedestrians, public roads, and highways, and assigns a unique identification number to each instance. We adhere to the official split protocols, dividing these videos into 15 for training and 3 for testing. The training set includes 80 distinct language descriptions, while the testing set includes 63.

\subsection{Evaluation Metrics}
We utilize Higher Order Tracking Accuracy (HOTA)~\cite{luiten2021hota} as the primary metric to evaluate the performance of our technique. Also, for further analysis, we include MOTA~\cite{bernardin2008clearmot} and IDF1~\cite{ristani2016idf1} as supplementary metrics.

\subsection{Implementation Details}

\textbf{Data Preprocessing}. The input data is categorized into three types: bounding boxes of tracked objects, video frames, and textual captions. Frames are cropped into images of objects (local images) through dataset bounding boxes. These cropped patches are then square-padded and resized into the shape of 224x224. The video frames are also square padded and resized to the shape of 672x672 (global images), and textual features are standardized using iKUN's~\cite{du2024ikun} vocabularies before being tokenized with the CLIP tokenizer~\cite{radford2021clip}.

\textbf{Network Architectures}.
For all experiments, we employed the SwinV2-T model \cite{liu2022swin}, pre-trained on the ImageNet-1k dataset \cite{deng2009imagenet}, as the encoder for both local and global images. Visual features were dimensionally reduced from $[n,16,768]$ to $[n,256]$ using multi-layer perceptrons (MLPs). The tokenized caption is fed into the original text encoder of CLIP-RN50 \cite{radford2021clip}, resulting in the encoded textual features. Textual features were encoded using CLIP-RN50 \cite{radford2021clip}, resulting in features of dimension $[n, 20, 1024]$. The final fusion features underwent spatio-temporal pooling using average pooling followed by max pooling. Using cosine similarity as the loss function, the model was trained for 100 epochs on ground-truth tracklets and relevant language expressions, while both the textual and visual encoders were frozen. We employed a similarity calibration method described in \cite{du2024ikun} without changes during evaluation.

\textbf{Model Configurations}.
During training, a batch size of 8 was used with a starting learning rate of 1e-5 and momentum of 1e-5. The training was conducted on a machine equipped with an \textit{Intel Xeon Processor E5 v3 Family @2.50 GHz, 48 GB RAM with P40} (P40 machine for short). 

During testing, a batch size of 1 was used on the same machine. For inference, we utilized MeMOTR \cite{MeMOTR} with the BDD100K checkpoint, named MeMOTR-BDD100K, for real-time tracking. Candidates predicted objects are then filtered with a score threshold of 0, using a machine with \textit{Intel Core i5-9300H @ 2.40GHz, 16 GB RAM with GTX1650} (GTX1650 machine for short).

\subsection{Evaluation Results}
Our method was evaluated against the state-of-the-art on Refer-KITTI using NeuralSORT~\cite{du2024ikun} as the base tracker and YOLOv8~\cite{Jocher_Ultralytics_YOLO_2023} as the detector. It is important to note that the tracking results are provided beforehand. Our approach, MEX, achieved 45.07\%, 32.81\%, and 62.52\% for HOTA, DetA, and AssA, respectively, surpassing the baseline method's results of 44.56\%, 32.05\%, and 62.49\%. This improvement demonstrates the efficacy of our approach while leveraging the advantages of iKUN~\cite{du2024ikun} with single-training efficiency.

\begin{figure}
    \centering
    \includegraphics[width=\linewidth]{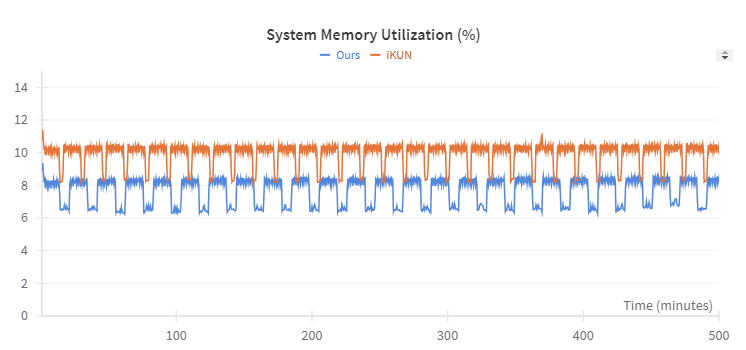}
    \includegraphics[width=\linewidth]{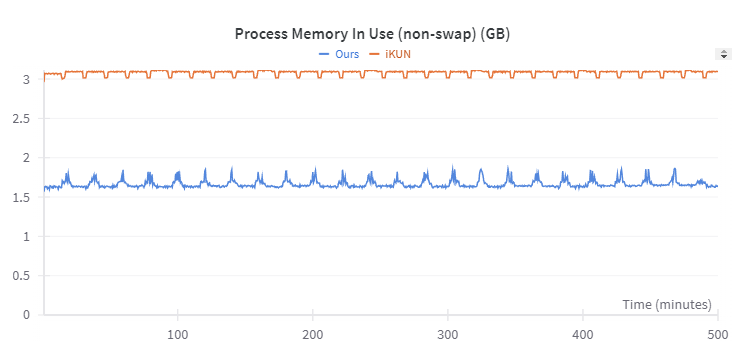}
    \caption{The system memory and process memory usage differences between iKUN's and ours when running training phase on the P40 machine.}
    \label{fig:mem_sys_process}
\end{figure}

\begin{table}[b]
    \scriptsize
    \caption{
    Comparison between iKUN and our approach on Refer-KITTI~\cite{wu2023referring}. Two referring modules are experimented with based on the results from NeuralSORT~\cite{du2024ikun}. Note that this analysis is based on the tracking results using YOLOv8~\cite{Jocher_Ultralytics_YOLO_2023} as the base detector in NeuralSORT. 
    }
    \centering
    \begin{tabularx}{\columnwidth}{|c|*{9}{X|}}
    \hline
        Module & HOTA &DetA &AssA &DetRe &DetPr &AssRe &MOTA &IDF1  \\\hline
        iKUN's &44.56 &32.05 &62.49 &48.53 &\textbf{44.76} &70.52 &\textbf{9.69} &55.40 \\\hline
        Ours &\textbf{45.07} &\textbf{32.81} &\textbf{62.52} & \textbf{54.84} &41.65 &\textbf{71.09} &1.82 & \textbf{56.52} \\\hline
    \end{tabularx}
    
    \label{tab:hota}
\end{table}
We also compare the number of trainable parameters counted in Table~\ref{tab:param}. Our referring module, MEX, has 81 million trainable parameters, less than iKUN~\cite{du2024ikun} with 92 million. With 10 million fewer trainable parameters, we can minimize memory usage during training. As a result, during training, our process memory utilization is significantly lower than iKUN's, approximately 47.20\% with system memory roughly 2\% for on the P40 machine. Fig.~\ref{fig:mem_sys_process} illustrates the memory allocation during training.

\begin{table}[t]
\caption{Comparison between iKUN's and MEX's number of trainable parameters in the referring module. Our technique decreases the number by over 10 million.}
    \centering
    \begin{tabular}{|c|c|}
    \hline
    Module & Number of trainable parameters \\ 
    \hline
    iKUN  & 92M  \\
    \hline
    MEX & \textbf{81M}  \\\hline
    \end{tabular}
    \label{tab:param}
\end{table}

\begin{figure}
    \centering
    \includegraphics[width=\linewidth]{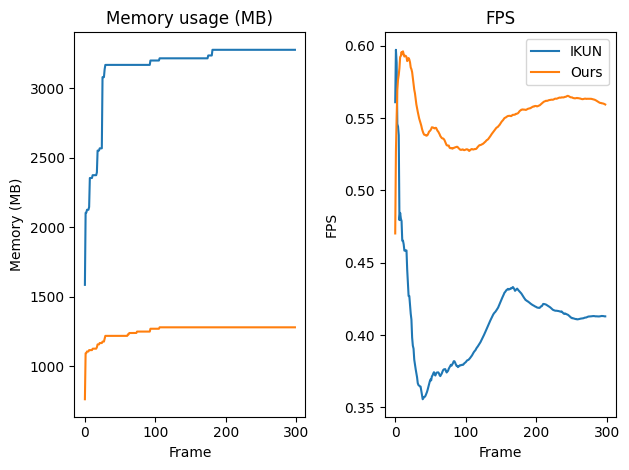}
    \caption{The memory usage and frame-per-second (FPS) differences when running inference with the MeMOTR-BDD100K~\cite{MeMOTR} tracker between iKUN's and ours. This experiment was conducted on the GTX1650 machine using a sequence from the Refer-KITTI testing set. Notably, the sequence includes cars and pedestrians.}
    \label{fig:mem_fps}
\end{figure}

Moreover, experiments were conducted using the MeMOTR-BDD100K~\cite{MeMOTR} tracker on Refer-KITTI's~\cite{wu2023referring} test set, running on the GTX1650 machine. During inference, we compared our memory usage and frame-per-second (FPS) with iKUN's \cite{du2024ikun}. iKUN consistently requires significantly more memory, between 954 and 2000 MB, compared to our approach.
Our method achieves an increased FPS of 0.15 to 0.2 frames and superior memory efficiency during real-time object tracking, as shown in Fig.~\ref{fig:mem_fps}. During inference, MeMOTR-BDD100K~\cite{MeMOTR} is observed to track cars and pedestrians effectively.

\subsection{Limitations}

\begin{figure}
    \centering
    \subfloat[people riding on bicycle]{
        \includegraphics[width=.4\linewidth]{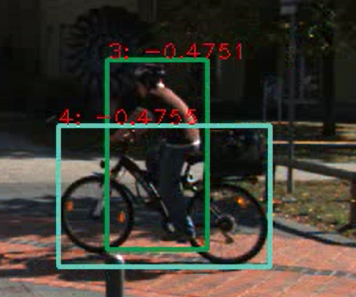}
    }\\
    \subfloat[people walking on the street]{
        \includegraphics[width=.4\linewidth]{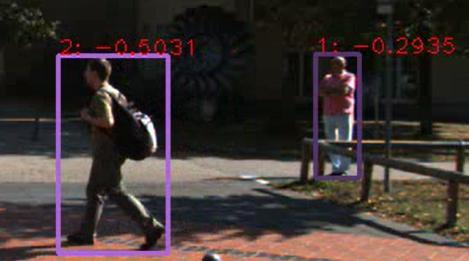}
    \includegraphics[width=.43\linewidth]{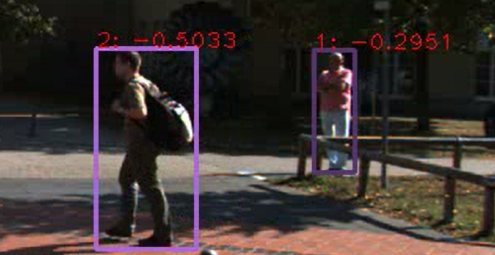}
    }
    \caption{Typical limitations in our pipeline. Best viewed in color.}
    \label{fig:limit}
\end{figure}

\textbf{Tracker}. Leveraging the MeMOTR-BDD100K~\cite{MeMOTR} tracker, our model demonstrates solid object identification capabilities but struggles with generalizing to complex objects. For instance, when querying ``people riding on bicycle'', as depicted in Fig.~\ref{fig:limit}(a), the tracker identifies two separate objects, the person and the bicycle, rather than recognizing them as a single entity. Consequently, our model evaluates these as distinct objects and fails to match the query accurately.

\textbf{Motion query}. Due to a lack of diverse motion description datasets for tracking, our model faces limitations in filtering objects based on motion queries. For example, when queried with ``people walking the street'', as witnessed in Fig.~\ref{fig:limit}(b), the model exhibits confusion between detecting walking or standing postures.

\section{Conclusion}

In this paper, we have introduced a novel approach to the referring module in the tracking-then-referring pipeline, solving the referring multi-object tracking task. Our method, the Memory-Efficient Cross-Modality (MEX) module, accommodates three types of necessary data sources for the proposed task, effectively addressing the cumbersome fusion steps between different modalities. This module helps the network work fine on the Refer-KITTI benchmark compared to the previous method, iKUN. Also, with the utilization of a more robust encoder, our designated pipeline works more efficiently, lessening memory usage and inference speed. 

\section*{Acknowledgments}

This research is supported by research funding from Faculty of Information Technology,
University of Science, Vietnam National University - Ho Chi Minh City.

\balance
\bibliographystyle{IEEEtran}
\nocite{*}
\bibliography{biblio}

\begin{thebibliography}{10}
\providecommand{\url}[1]{#1}
\csname url@samestyle\endcsname
\providecommand{\newblock}{\relax}
\providecommand{\bibinfo}[2]{#2}
\providecommand{\BIBentrySTDinterwordspacing}{\spaceskip=0pt\relax}
\providecommand{\BIBentryALTinterwordstretchfactor}{4}
\providecommand{\BIBentryALTinterwordspacing}{\spaceskip=\fontdimen2\font plus
\BIBentryALTinterwordstretchfactor\fontdimen3\font minus \fontdimen4\font\relax}
\providecommand{\BIBforeignlanguage}[2]{{%
\expandafter\ifx\csname l@#1\endcsname\relax
\typeout{** WARNING: IEEEtran.bst: No hyphenation pattern has been}%
\typeout{** loaded for the language `#1'. Using the pattern for}%
\typeout{** the default language instead.}%
\else
\language=\csname l@#1\endcsname
\fi
#2}}
\providecommand{\BIBdecl}{\relax}
\BIBdecl

\bibitem{wu2023referring}
D.~Wu, W.~Han, T.~Wang, X.~Dong, X.~Zhang, and J.~Shen, ``Referring multi-object tracking,'' in \emph{Proceedings of the IEEE/CVF Conference on Computer Vision and Pattern Recognition}, 2023, pp. 14\,633--14\,642.

\bibitem{zeng2021motr}
F.~Zeng, B.~Dong, Y.~Zhang, T.~Wang, X.~Zhang, and Y.~Wei, ``Motr: End-to-end multiple-object tracking with transformer,'' in \emph{European Conference on Computer Vision (ECCV)}, 2022.

\bibitem{nguyen2023type}
P.~Nguyen, K.~G. Quach, K.~Kitani, and K.~Luu, ``Type-to-track: Retrieve any object via prompt-based tracking,'' \emph{Advances in Neural Information Processing Systems}, 2023.

\bibitem{du2024ikun}
Y.~Du, C.~Lei, Z.~Zhao, and F.~Su, ``ikun: Speak to trackers without retraining,'' in \emph{Proceedings of the IEEE/CVF Conference on Computer Vision and Pattern Recognition (CVPR)}, June 2024, pp. 19\,135--19\,144.

\bibitem{liu2022swin}
Z.~Liu, H.~Hu, Y.~Lin, Z.~Yao, Z.~Xie, Y.~Wei, J.~Ning, Y.~Cao, Z.~Zhang, L.~Dong \emph{et~al.}, ``Swin transformer v2: Scaling up capacity and resolution,'' in \emph{Proceedings of the IEEE/CVF conference on computer vision and pattern recognition}, 2022, pp. 12\,009--12\,019.

\bibitem{radford2021clip}
A.~Radford, J.~W. Kim, C.~Hallacy, A.~Ramesh, G.~Goh, S.~Agarwal, G.~Sastry, A.~Askell, P.~Mishkin, J.~Clark \emph{et~al.}, ``Learning transferable visual models from natural language supervision,'' in \emph{International conference on machine learning}.\hskip 1em plus 0.5em minus 0.4em\relax PMLR, 2021, pp. 8748--8763.

\bibitem{luiten2021hota}
J.~Luiten, A.~Osep, P.~Dendorfer, P.~Torr, A.~Geiger, L.~Leal-Taix{\'e}, and B.~Leibe, ``Hota: A higher order metric for evaluating multi-object tracking,'' \emph{International journal of computer vision}, vol. 129, pp. 548--578, 2021.

\bibitem{wojke2017simple}
N.~Wojke, A.~Bewley, and D.~Paulus, ``Simple online and realtime tracking with a deep association metric,'' in \emph{2017 IEEE international conference on image processing (ICIP)}.\hskip 1em plus 0.5em minus 0.4em\relax IEEE, 2017, pp. 3645--3649.

\bibitem{Wojke2018deepsort}
N.~Wojke and A.~Bewley, ``Deep cosine metric learning for person re-identification,'' in \emph{2018 IEEE Winter Conference on Applications of Computer Vision (WACV)}.\hskip 1em plus 0.5em minus 0.4em\relax IEEE, 2018, pp. 748--756.

\bibitem{zhang2022bytetrack}
Y.~Zhang, P.~Sun, Y.~Jiang, D.~Yu, F.~Weng, Z.~Yuan, P.~Luo, W.~Liu, and X.~Wang, ``Bytetrack: Multi-object tracking by associating every detection box,'' in \emph{European conference on computer vision}.\hskip 1em plus 0.5em minus 0.4em\relax Springer, 2022, pp. 1--21.

\bibitem{aharon2022botsort}
N.~Aharon, R.~Orfaig, and B.-Z. Bobrovsky, ``Bot-sort: Robust associations multi-pedestrian tracking,'' \emph{arXiv preprint arXiv:2206.14651}, 2022.

\bibitem{cao2023observation}
J.~Cao, J.~Pang, X.~Weng, R.~Khirodkar, and K.~Kitani, ``Observation-centric sort: Rethinking sort for robust multi-object tracking,'' in \emph{Proceedings of the IEEE/CVF Conference on Computer Vision and Pattern Recognition}, 2023, pp. 9686--9696.

\bibitem{vaswani2017attention}
A.~Vaswani, N.~Shazeer, N.~Parmar, J.~Uszkoreit, L.~Jones, A.~N. Gomez, {\L}.~Kaiser, and I.~Polosukhin, ``Attention is all you need,'' \emph{Advances in neural information processing systems}, vol.~30, 2017.

\bibitem{meinhardt2021trackformer}
T.~Meinhardt, A.~Kirillov, L.~Leal-Taixe, and C.~Feichtenhofer, ``Trackformer: Multi-object tracking with transformers,'' in \emph{The IEEE Conference on Computer Vision and Pattern Recognition (CVPR)}, June 2022.

\bibitem{zhang2023motrv2}
Y.~Zhang, T.~Wang, and X.~Zhang, ``Motrv2: Bootstrapping end-to-end multi-object tracking by pretrained object detectors,'' in \emph{Proceedings of the IEEE/CVF Conference on Computer Vision and Pattern Recognition}, 2023, pp. 22\,056--22\,065.

\bibitem{yolox2021}
Z.~Ge, S.~Liu, F.~Wang, Z.~Li, and J.~Sun, ``Yolox: Exceeding yolo series in 2021,'' \emph{arXiv preprint arXiv:2107.08430}, 2021.

\bibitem{cai2022memot}
J.~Cai, M.~Xu, W.~Li, Y.~Xiong, W.~Xia, Z.~Tu, and S.~Soatto, ``Memot: Multi-object tracking with memory,'' in \emph{Proceedings of the IEEE/CVF Conference on Computer Vision and Pattern Recognition}, 2022, pp. 8090--8100.

\bibitem{MeMOTR}
R.~Gao and L.~Wang, ``{MeMOTR}: Long-term memory-augmented transformer for multi-object tracking,'' in \emph{Proceedings of the IEEE/CVF International Conference on Computer Vision (ICCV)}, October 2023, pp. 9901--9910.

\bibitem{ZHAO202310}
\BIBentryALTinterwordspacing
H.~Zhao, X.~Wang, D.~Wang, H.~Lu, and X.~Ruan, ``Transformer vision-language tracking via proxy token guided cross-modal fusion,'' \emph{Pattern Recognition Letters}, vol. 168, pp. 10--16, 2023. [Online]. Available: \url{https://www.sciencedirect.com/science/article/pii/S0167865523000545}
\BIBentrySTDinterwordspacing

\bibitem{zhou2023joint}
L.~Zhou, Z.~Zhou, K.~Mao, and Z.~He, ``Joint visual grounding and tracking with natural language specification,'' in \emph{Proceedings of the IEEE/CVF conference on computer vision and pattern recognition}, 2023, pp. 23\,151--23\,160.

\bibitem{zheng2023towards}
Y.~Zheng, B.~Zhong, Q.~Liang, G.~Li, R.~Ji, and X.~Li, ``Towards unified token learning for vision-language tracking,'' \emph{IEEE Transactions on Circuits and Systems for Video Technology}, 2023.

\bibitem{wu2022language}
J.~Wu, Y.~Jiang, P.~Sun, Z.~Yuan, and P.~Luo, ``Language as queries for referring video object segmentation,'' in \emph{Proceedings of the IEEE/CVF Conference on Computer Vision and Pattern Recognition}, 2022, pp. 4974--4984.

\bibitem{botach2021end}
A.~Botach, E.~Zheltonozhskii, and C.~Baskin, ``End-to-end referring video object segmentation with multimodal transformers,'' in \emph{Proc. IEEE Conf. Computer Vision and Pattern Recognition (CVPR)}, 2022.

\bibitem{Geiger2012CVPR}
A.~Geiger, P.~Lenz, and R.~Urtasun, ``Are we ready for autonomous driving? the kitti vision benchmark suite,'' in \emph{Conference on Computer Vision and Pattern Recognition (CVPR)}, 2012.

\bibitem{bernardin2008clearmot}
K.~Bernardin and R.~Stiefelhagen, ``Evaluating multiple object tracking performance: the clear mot metrics,'' \emph{EURASIP Journal on Image and Video Processing}, vol. 2008, pp. 1--10, 2008.

\bibitem{ristani2016idf1}
E.~Ristani, F.~Solera, R.~Zou, R.~Cucchiara, and C.~Tomasi, ``Performance measures and a data set for multi-target, multi-camera tracking,'' in \emph{European conference on computer vision}.\hskip 1em plus 0.5em minus 0.4em\relax Springer, 2016, pp. 17--35.

\bibitem{deng2009imagenet}
J.~Deng, W.~Dong, R.~Socher, L.-J. Li, K.~Li, and L.~Fei-Fei, ``Imagenet: A large-scale hierarchical image database,'' in \emph{2009 IEEE conference on computer vision and pattern recognition}.\hskip 1em plus 0.5em minus 0.4em\relax Ieee, 2009, pp. 248--255.

\bibitem{Jocher_Ultralytics_YOLO_2023}
\BIBentryALTinterwordspacing
G.~Jocher, A.~Chaurasia, and J.~Qiu, ``{Ultralytics YOLO},'' Jan. 2023. [Online]. Available: \url{https://github.com/ultralytics/ultralytics}
\BIBentrySTDinterwordspacing

\bibitem{dendorfer2021motchallenge}
P.~Dendorfer, A.~Osep, A.~Milan, K.~Schindler, D.~Cremers, I.~Reid, S.~Roth, and L.~Leal-Taix{\'e}, ``Motchallenge: A benchmark for single-camera multiple target tracking,'' \emph{International Journal of Computer Vision}, vol. 129, pp. 845--881, 2021.

\bibitem{focalloss}
T.-Y. Lin, P.~Goyal, R.~Girshick, K.~He, and P.~Dollár, ``Focal loss for dense object detection,'' in \emph{2017 IEEE International Conference on Computer Vision (ICCV)}, 2017, pp. 2999--3007.

\end{thebibliography}

\end{document}